\documentclass[11pt]{article}

\usepackage[final]{acl}

\usepackage{times}
\usepackage{latexsym}
\usepackage[T1]{fontenc}
\usepackage[utf8]{inputenc}
\usepackage{microtype}
\usepackage{inconsolata}
\usepackage{graphicx}
\usepackage{placeins}
\usepackage{listings}
\usepackage{xcolor}
\usepackage{caption}
\usepackage{subcaption}
\usepackage{booktabs}
\usepackage{enumitem}
\setlist{itemsep=2pt,topsep=3pt,parsep=0pt}
\usepackage{needspace}
\title{CARF: Context-Aware Reasoning Supervision for Scientific Evidence Screening}

\author{
  \textbf{Nishtha Agarwal\textsuperscript{1}}\thanks{Lead author.} \quad
  \textbf{Adam Visokay\textsuperscript{2}} \quad
  \textbf{Astha Wahi\textsuperscript{3}} \quad
  \textbf{Kwabena Owusu Aninkora\textsuperscript{3,4}}
  \\
  \textbf{Amy Gu\textsuperscript{5}} \quad
  \textbf{Tyler H. McCormick\textsuperscript{1}} \quad
  \textbf{Li Liu\textsuperscript{3}}
  \\
  \textsuperscript{1}University of Washington \quad
  \textsuperscript{2}University of California, Berkeley \quad
  \textsuperscript{3}Johns Hopkins University
  \\
  \textsuperscript{4}SUNY Downstate Health Sciences University \quad
  \textsuperscript{5}Independent Researcher
  \\[2pt]
  {\footnotesize\texttt{nishtha1@uw.edu, avisokay@ischool.berkeley.edu, awahi1@jhmi.edu,}}
  \\
  {\footnotesize\texttt{kaninko1@jh.edu, guamy56@gmail.com, tylermc@uw.edu, lliu26@jhu.edu}}
}

\begin{document}
\maketitle

\begin{abstract}
Systematic reviews synthesize biomedical evidence to inform clinical and public-health decision-making, but require domain experts to manually screen thousands of studies against detailed eligibility criteria. We investigate whether language models can assist with this bottleneck by learning from expert feedback on title and abstract screening decisions and model-generated reasoning. We introduce the Context-Aware Reasoning Framework (CARF), a human-in-the-loop framework that uses expert feedback on model-generated reasoning to adapt language models to context-dependent screening tasks. We evaluate CARF on systematic reviews of verbal autopsy and cause-of-death ascertainment methods used in global public health mortality surveillance. CARF shows competitive screening performance compared to evaluated models while retaining high sensitivity, and is more inclined to retain ambiguous, context-dependent studies for human review. Together, these findings suggest that expert-corrected reasoning is a promising supervision signal for context-dependent scientific screening, while retaining a role for human oversight.
\end{abstract}

\section{Introduction}

Systematic reviews are central to evidence-based medicine and public health, synthesizing findings from diverse studies to guide clinical practice, biomedical research, and public health policy. However, producing a systematic review remains a labor-intensive and time-consuming process \citep{toth2024}, requiring domain experts to screen thousands or more publications for eligibility and extract nuanced evidence from full-text articles. The process typically begins with comprehensive literature searches to identify potentially relevant studies, followed by title and abstract screening to exclude records that do not meet predefined eligibility criteria. Studies retained after this initial screening undergo full-text assessment to determine final eligibility, followed by data extraction and synthesis of the included evidence. Title and abstract screening is therefore a critical early stage, as its decisions determine which studies proceed to more resource-intensive assessment. These decisions often depend on interpreting complex clinical language and study-specific context, involving subjective judgments, making the process difficult to automate \citep{mahmoudi2025}.

The need for scalable evidence synthesis is particularly acute in global health, where data tend to be sparse and evidence must be aggregated across geographic and temporal contexts. In this paper, we focus on title and abstract screening for verbal autopsy (VA) and cause-of-death (COD) ascertainment. Verbal autopsy is a widely used approach for estimating causes of death in regions where deaths occur outside healthcare facilities and medical certification is unavailable or incomplete \citep{who2022}. It relies on structured interviews with caregivers or family members containing both structured and free-text information about the signs, symptoms, and circumstances preceding death, from which a probable cause of death is assigned by physicians or automated algorithms \citep{who2022}. As new VA questionnaires, physician review protocols, and machine learning--based COD assignment methods continue to emerge \citep{yoshida2023bayesian, fan2024narratives_colm, chen2025lava}, systematic reviews are essential for comparing their performance and synthesizing evidence across diverse populations and settings to inform global mortality surveillance standards \citep{mahesh2023}. The diversity of study designs, populations, and COD assignment approaches makes consistent eligibility assessment particularly challenging, motivating the use of large language models (LLMs) to assist experts with screening and evidence extraction while preserving the rigor of systematic reviews.

Despite their strong language reasoning capabilities, LLMs remain limited by the supervision available for systematic review tasks \citep{anisuzzaman2025}. Existing systematic review datasets typically provide only final screening decisions or extracted variables, without the reasoning, disagreements, and contextual considerations that underpin expert assessments \citep{akinseloyin2024, guo2024}. As a result, language models can learn to reproduce inclusion or exclusion labels but receive little supervision for learning how experts interpret ambiguous evidence, apply domain-specific criteria, and resolve context-dependent eligibility decisions. To address this gap, we investigate whether expert reasoning can serve as an effective supervision signal for better aligning language-model decisions with scientific judgment in context-dependent systematic review screening tasks.

 To this end, we introduce the Context-Aware Reasoning Framework (CARF), a human-in-the-loop framework that converts expert natural-language explanations into supervision for language-model adaptation. Starting with a set of screened articles by domain experts, a reasoning model screens each article blind and produces chain-of-thought reasoning with an include/exclude decision from the title and abstract, without access to the reviewer's label. Domain experts then review the cases where this decision and reasoning diverges from the verified decision and provide targeted natural-language corrections and feedback. This is then paired with the model's original (flawed) reasoning, to construct a dataset of expert-corrected reasoning. A smaller reasoning model is then adapted on this dataset through parameter-efficient Low-Rank Adaptation (LoRA) fine-tuning, learning not only from inclusion/exclusion labels but from the expert deliberation and reasoning itself.

Our contributions are threefold. First, we introduce CARF, a human-in-the-loop framework for constructing expert reasoning supervision for context-dependent systematic review screening.
Second, we show that CARF yields better alignment with expert screening decisions than common alternatives. Third, we develop a domain-adapted reasoning model and a dataset with expert reasoning for verbal autopsy and cause-of-death systematic review screening, providing resources to support future research in biomedical evidence synthesis.

\section{Related Work}

\subsection{Traditional LLMs for Systematic Review Screening and Evidence Extraction}

Large language model (LLM)-based approaches have demonstrated promising performance on title and abstract screening and structured evidence extraction through prompting-based methods \citep{helmsandersen2025, sercombe2025}. \citet{parmar2026} further explored collaborative inference strategies that combine predictions from multiple language models to improve screening reliability. Although these approaches substantially reduce reviewer workload and improve screening performance, they continue to require human oversight and validation before deployment, indicating that reliably reproducing expert screening decisions remains an open challenge.

Beyond screening, studies have also explored the use of LLMs for automated data extraction from full-text articles. GPT-based systems have been evaluated for extracting predefined study characteristics and other structured variables required for systematic reviews, demonstrating the potential of LLMs to accelerate evidence extraction while still requiring human validation to ensure accuracy and consistency \citep{gartlehner2024, kataoka2026, mahmoudi2025}. Notably, \citet{mahmoudi2025} found that LLMs performed well on extracting explicitly stated information but showed substantially lower accuracy on nuanced, subjective criteria requiring contextual interpretation, underscoring the continued need for human oversight in complex evidence synthesis tasks.

\subsection{Adapting Language Models to Systematic Review Tasks}

Bio-SIEVE introduced instruction tuning for systematic review automation \citep{robinson2023}, and \citet{yamoah2026} investigated supervised fine-tuning of an open-weight language model for title and abstract screening, showing that task-specific adaptation can improve screening performance compared with prompting and base models. These approaches primarily supervise models using final screening outcomes. Bio-SIEVE additionally explored generating exclusion reasons as an auxiliary task, using brief post-hoc justifications recorded by Cochrane reviewers. However, this multi-task training reduced screening performance relative to label-only training, which the authors attributed partly to justifications that depended on full-text information unavailable in the abstract \citep{robinson2023}. Neither approach supervises the reasoning process that leads to a decision, or uses expert corrections of a model's own reasoning to capture how reviewers interpret ambiguous evidence and resolve context-dependent eligibility decisions.

\subsection{Reasoning Supervision for Language Models}

Chain-of-thought prompting demonstrated that generating intermediate reasoning can substantially improve language model performance on multi-step reasoning tasks \citep{wei2022}. Building on this idea, subsequent work leveraged reasoning traces as supervision by bootstrapping models through self-generated rationales, distilling teacher-generated reasoning into smaller models, and incorporating process supervision during fine-tuning \citep{zelikman2022, hsieh2023, jiang2025}. These approaches are typically trained on question--reasoning--answer triples for tasks such as mathematical reasoning, commonsense reasoning, question answering, and code generation. Researchers have also explored improving reasoning through human correction of intermediate reasoning steps \citep{cai2023}. Further, there is evidence that language models struggle to reliably identify flaws in their own reasoning but can improve significantly when provided with explicit error localization or corrective feedback \citep{tyen2024}.

\begin{figure*}[t]
  \centering
  \includegraphics[width=\textwidth,keepaspectratio]{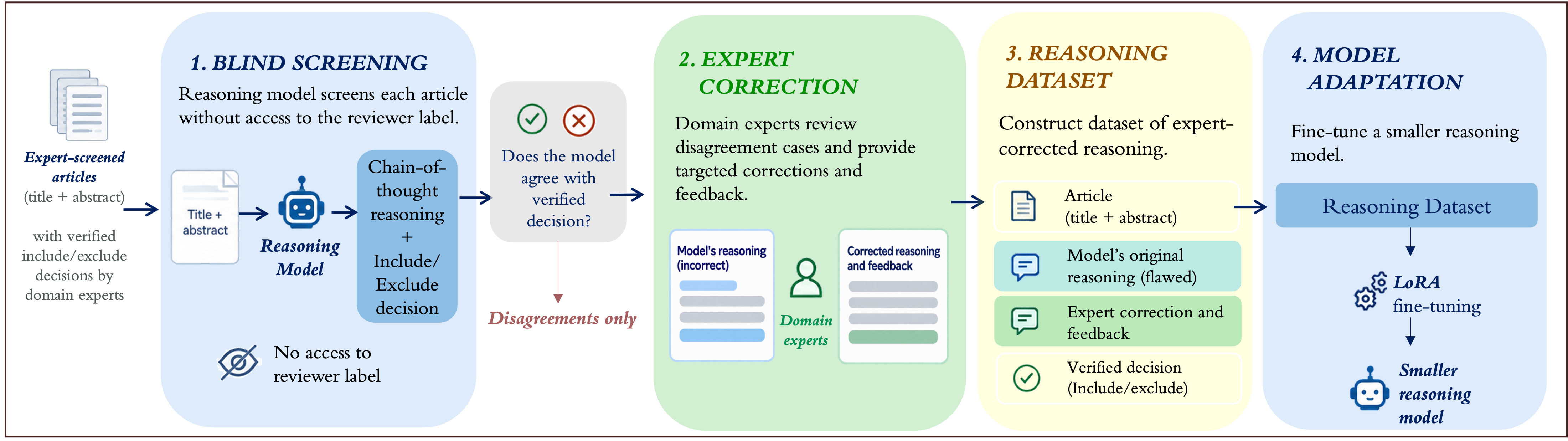}
  \caption{Overview of the Context-Aware Reasoning Framework (CARF) applied to the title and abstract screening stage of a systematic review, illustrating four stages: blind model screening, expert correction of disagreements, construction of an expert-reasoned dataset, and reasoning-based model adaptation through LoRA fine-tuning.}
  \label{fig:carf_overview}
\end{figure*}

\section{Methods}

We propose the Context-Aware Reasoning Framework (CARF), a four-stage framework for constructing expert reasoning supervision and adapting reasoning language models to context-dependent scientific decision-making. CARF uses verified expert screening decisions and targeted expert feedback on model reasoning to provide supervision for reasoning model adaptation. The framework consists of four stages: (1) literature retrieval and expert annotation of a subset of articles, (2) blind screening by a reasoning model using article titles and abstracts, (3) construction of an expert-corrected reasoning dataset from cases where model decisions diverge from verified expert decisions, and (4) reasoning-supervised language model adaptation through parameter-efficient LoRA fine-tuning. Figure~\ref{fig:carf_overview} illustrates the overall framework.

\subsection{Literature Retrieval and Expert Annotation}

Candidate studies are retrieved from relevant scientific literature databases using domain-specific search strategies. Three domain experts independently review the retrieved articles, with each article reviewed by only one expert for title and abstract screening according to predefined eligibility criteria. Articles are distributed across the three experts such that the full screening corpus is covered without requiring multiple experts to review the same article. 

Following this initial annotation, we take a first pass at blind screening outside the CARF framework. A reasoning model screens the same articles without access to the original expert decisions. Articles for which the model's decision disagrees with the expert annotation are subsequently returned to the human reviewers for further assessment. This human-in-the-loop design limits the need for exhaustive multi-reviewer screening, thereby reducing the time and resources required for large-scale screening. The approach is consistent with prior work advocating researcher-in-the-loop use of AI for systematic review screening to accelerate the review process while maintaining methodological oversight \citep{vanDijk2023AI}. 

The resulting expert decisions serve as the reference annotations for subsequent model screening and reasoning supervision.

\subsection{Blind Screening}
\label{sec:blind-screening}

Following expert annotation and verification, CARF now conducts a blind screening stage to obtain an independent model assessment of each article. A reasoning language model (our teacher model) is provided only with the article title and abstract and is not given access to the expert screening decision. The model is prompted to determine whether the article should be included or excluded according to the systematic review criteria and to produce a natural-language reasoning trace supporting its decision. The resulting model decision is then compared with the verified expert decision to identify cases of disagreement. Now, instead of treating these disagreements solely as errors to be adjudicated, CARF uses them to identify instances in which the model's reasoning fails to capture the contextual considerations reflected in expert judgment. These disagreement cases are subsequently passed to domain experts for targeted correction and feedback in the expert reasoning dataset construction stage.

\subsection{Expert Correction and Reasoning Dataset Construction}
\label{sec:expert-reasoning-dataset}

CARF constructs an expert reasoning supervision dataset from all blind-screened articles, with expert feedback added for cases where the model's decision diverges from the verified expert decision. For each disagreement case, domain experts review the article, the model's original reasoning, and the corresponding screening criteria. The experts then identify shortcomings in the model's reasoning and provide targeted natural-language corrections and contextual feedback. This feedback may clarify relevant evidence in the article, explain the application of eligibility criteria, or identify contextual considerations that led to the expert decision.

The expert feedback is paired with the model's original reasoning and the verified screening decision to construct reasoning supervision examples. Each disagreement example therefore captures the relationship between the article content, the model's initial reasoning, the expert's correction of the reasoning, and the verified expert decision. This process transforms model errors into targeted expert supervision, providing training signals that capture not only whether a screening decision is correct, but also how expert judgment differs from the model's initial reasoning. Articles on which the model's blind decision agreed with the expert are retained with their original reasoning and verified decision, without a feedback block.

The resulting expert reasoning dataset is used to adapt a smaller reasoning language model in the subsequent reasoning-supervised fine-tuning stage.

\subsection{Reasoning-Supervised Language Model Adaptation}

A smaller reasoning model (student) is adapted through parameter-efficient fine-tuning using Low-Rank Adaptation (LoRA) \citep{hu2022lora}, trained on the expert reasoning supervision dataset. LoRA freezes the base model weights and trains a small number of additional low-rank parameters, reducing computational cost and mitigating overfitting \citep{hu2022lora}. This supports the scale of our expert-verified dataset. By learning from both the teacher-generated reasoning and expert feedback, the adapted model learns the contextual decision boundaries underlying systematic review screening beyond the final inclusion or exclusion label.

\FloatBarrier

\section{Experiments}


\subsection{Literature Retrieval}

We constructed our systematic review corpus by searching five scientific literature databases: PubMed, Embase, Scopus, the ACM Digital Library, and arXiv. Search strategies were designed to retrieve studies at the intersection of verbal autopsy, cause-of-death ascertainment, and use of artificial intelligence, incorporating keyword or MeSH terms based queries. Search queries were adapted to the syntax of each database while consistently targeting three concepts: (1) verbal autopsy and cause-of-death ascertainment, (2) mortality and death, and (3) artificial intelligence and computational methods. The complete search strategies are provided in Appendix~\ref{app:search}.

\subsection{Screening Dataset Construction}

Screening decisions were exported from DistillerSR using the title and abstract screening form. The exported dataset contained 5,263 screening records but did not include article abstracts. Each record was assigned one of six screening outcomes: 1) Include, 2) Exclude: No Cause-of-Death Data, 3) Exclude: Fewer than 15 Deaths, 4) Exclude: Missing Abstract/Full Text, 5) Exclude: Non-Human, and 6) Exclude: Duplicate. For studies meeting the inclusion criteria, reviewers additionally completed ten structured eligibility fields describing study characteristics, including language, study population, number of causes of death, and the use of AI-based methods.

Article abstracts were retrieved separately through the DistillerSR REST API and merged with the exported screening records using the reference identifier (RefID). Metadata were successfully retrieved for all 5,263 screening records, and abstracts were recovered for 5,161 records (98.06\%), including all included studies. The expert labeled corpus comprised 1,072 included studies (20.8\%), 3,838 studies excluded for lacking cause-of-death data (74.4\%), 144 studies excluded for reporting fewer than 15 deaths (2.8\%), 3 studies excluded due to missing abstract or full text (0.06\%), 71 non-human studies (1.4\%), and 33 duplicate records (0.6\%). These numbers reflect the post-adjudication results finalized through the verification process described in Section~\ref{sec:verification}.

\subsection{Human Annotation}

Title and abstract screening was performed by three domain experts following predefined eligibility criteria. Each article was assigned to a single reviewer, who determined whether the study met the review inclusion criteria and selected one of the six screening decisions within DistillerSR. For studies meeting the inclusion criteria, reviewers additionally completed the structured eligibility fields used during downstream evidence synthesis. The complete eligibility criteria is provided in Appendix~\ref{app:screening}.

\subsection{Model-Assisted Ground-Truth Verification}
\label{sec:verification}

We employed DeepSeek-R1-Distill-Qwen-32B as the screening model that received only the article title and abstract, without access to the original human annotation. The model assigned one of the screening decisions to each article. 
The model's predictions were compared with the original reviewer annotations for all 5,263 screened articles. Each comparison was categorized into one of several agreement types, including exact agreement, disagreement in screening direction (include versus exclude), disagreement in exclusion reason, and model uncertainty. Across the corpus, the model agreed with the original reviewer on 3,139 articles ($\sim$60\%), while 2,104 articles ($\sim$40\%) were flagged for adjudication based on disagreement; the model did not reach a sufficient decision on the remaining 20 articles.
To assess whether disagreement corresponded to genuine annotation issues and not random model errors, we evaluated this blind-screening procedure on a small set of previously identified annotation errors. The model successfully identified 11 of 14 known labeling errors (78.6\%), suggesting that model disagreement provides a useful signal for prioritizing expert review. The complete prompt used for this stage is provided in Appendix~\ref{app:prompt-verification}.

\subsection{Experimental Data Splits}

The 5,161 retrieved articles were partitioned into training, validation, and held-out test sets to preserve the distribution of screening decisions. A total of 1,000 articles were reserved exclusively for testing and were not used during model fine-tuning. Of the remaining articles, 300 were held out as a validation set for checkpoint selection during model adaptation, while the remaining 3,861 articles were used as a training set for expert correction and reasoning dataset construction described in section~\ref{sec:expert-reasoning-dataset} and model training.

\subsection{Expert Reasoning Supervision Dataset Construction}

Because CARF uses natural-language reasoning as the supervision signal, the teacher model must generate explicit reasoning traces that can be reviewed and refined by experts. We therefore again employ the reasoning-oriented DeepSeek-R1-Distill-Qwen-32B model, because models in this family have demonstrated strong reasoning performance in prior work \citep{moell2025, chua2025, zhan2025}, to generate screening decisions and rationales for those decisions for all 3,861 training articles. We select an open-weight reasoning model to ensure reproducibility and enable scalable generation of reasoning traces. Of the 3,861 articles, 3 were dropped during preprocessing and 25 did not yield a usable blinded decision: 22 were skipped as duplicates or missing text, 2 produced unparseable output, and 1 was truncated. Across the remaining 3,833 screened articles, the blind model's decision matched the verified reviewer label for 2,487 articles and disagreed for 1,346 articles.
One expert reviews each disagreement case and examines the model-generated reasoning alongside the article title, abstract, and screening criteria. For each case, expert provides targeted natural-language corrections and feedback addressing incomplete, incorrect, or unsupported reasoning and clarifying the contextual considerations underlying the verified screening decision. Expert feedback is attached to all 1,346 disagreement traces, while 114 cases additionally result in a change from the original screening decision. The resulting set of model-generated reasoning, verified screening decisions, and expert corrections forms the expert reasoning supervision dataset used for reasoning model adaptation. The instructions provided to the model for blind screening and generating reasoning traces is provided in Appendix~\ref{app:prompt-blind}. 

\subsection{Model Adaptation}

We adapt DeepSeek-R1-Distill-Qwen-14B as the student model using Low-Rank Adaptation (LoRA) \citep{hu2022lora}. Each training instance consists of the systematic-review screening instructions, article title, and abstract as input, with the corresponding reasoning trace and screening decision as the target output. For instances where the model's blind reasoning was in error, the original trace is paired with the expert's natural-language feedback identifying the flaw; the remaining instances retain the model's blind reasoning without feedback. During training, loss is computed only over the target sequence, so supervision is derived exclusively from the target reasoning trace, expert feedback, and decision, never the input. LoRA adapters are applied to both the attention and feed-forward projection layers, following common practice for reasoning-oriented adaptation, while the pretrained model weights remain frozen. Hyperparameters are selected a priori based on commonly adopted LoRA configurations for instruction tuning and reasoning models and were fixed throughout all experiments. Models were trained for three epochs, and loss was computed on the 300-article validation set at the end of each epoch. The checkpoint with the lowest validation loss was retained. No test-set information was used for checkpoint selection. The complete set of training hyperparameters is provided in Appendix~\ref{app:impl}.

\subsection{Baselines and Comparison Methods}

We compare CARF against four baselines spanning conventional classification, fine-tuning without expert reasoning, and inference-time prompting: (1) a PubMedBERT encoder baseline \citep{gu2022}, fine-tuned end-to-end as a binary Include/Exclude classifier; (2) a label-only fine-tuning baseline that fine-tunes DeepSeek-R1-Distill-Qwen-14B to predict the screening decision alone, without reasoning traces or expert feedback; (3) a reasoning-only teacher-distillation baseline that fine-tunes the same model on the teacher's blind-screening reasoning traces with no human supervision, isolating the contribution of expert reasoning supervision; and (4) a zero-shot baseline evaluating three frontier reasoning models: DeepSeek-V3.2, Grok-4.3, and Claude Sonnet-5, under the same screening instructions without task-specific fine-tuning. All three fine-tuned models (label-only, teacher-distilled reasoning, and CARF) are trained on the same training split (3,858 articles for label-only and 3,833 for the two reasoning models, which require a usable blind-screening trace) and share the same prompt template and LoRA fine-tuning configuration (Appendix~\ref{app:baselines}, Table~\ref{tab:hparams}); the zero-shot models involve no training. The complete set of training prompt given to the models is provided in Appendix~\ref{app:prompt-training}.

\subsection{Evaluation Metrics}
 
Because systematic reviews prioritize identifying all relevant studies, recall (sensitivity) for the Include class serves as the primary evaluation metric: false negatives correspond to eligible studies incorrectly excluded, which may compromise the completeness of the review and are typically unrecoverable in later screening stages. False positives, by contrast, are corrected at low cost during full-text review. To characterize this trade-off we additionally report precision, F1-score, and specificity for the Include class. Because the screening task is highly class-imbalanced due to the large majority of articles being excluded,  we focus on Include-class metrics rather than overall accuracy, which is dominated by the majority class and can mask differences in the model's ability to identify eligible studies.

\section{Results}

\begin{table*}[t]
\centering
\small
\setlength{\tabcolsep}{6pt}
\begin{tabular}{lcccccc}
\toprule
\textbf{Method} & \textbf{Rec.} & \textbf{95\% CI} & \textbf{Spec.} & \textbf{Prec.} & \textbf{F1} & \textbf{FP} \\
\midrule
\multicolumn{7}{l}{\textit{Fine-tuned (ours, 14B)}} \\
CARF                       & \textbf{0.964} & 0.923--0.987 & 0.857 & 0.573 & \textbf{0.719} & 119 \\
Reasoning (teacher-distilled) & \textbf{0.964} & 0.923--0.987 & 0.610 & 0.330 & 0.492 & 325 \\
Label-only                 & 0.910 & 0.855--0.949 & 0.873 & 0.588 & 0.714 & 106 \\
\midrule
\multicolumn{7}{l}{\textit{Zero-shot frontier}} \\
DeepSeek-V3.2   & 0.940 & 0.892--0.971 & 0.821 & 0.511 & 0.662 & 149 \\
Grok-4.3        & 0.922 & 0.870--0.958 & 0.853 & 0.554 & 0.692 & 123 \\
Claude Sonnet-5 & 0.922 & 0.870--0.958 & 0.860 & 0.567 & 0.702 & 117 \\
\midrule
\multicolumn{7}{l}{\textit{Encoder baseline}} \\
PubMedBERT      & 0.729 & 0.655--0.795 & \textbf{0.933} & \textbf{0.684} & 0.706 & \textbf{56} \\
\bottomrule
\end{tabular}
\caption{Include-class screening performance on the held-out test set ($n{=}1{,}000$; 166 eligible studies). CARF achieves the most favorable combination of recall (0.964) and F1 score (0.719) among all evaluated methods, outperforming zero-shot frontier models on both metrics. Fine-tuned results are from a single training run; zero-shot results are from a single inference pass. 95\% confidence intervals for recall use the Clopper–Pearson method.}
\label{tab:screening-results}
\end{table*}

As shown in Table~\ref{tab:screening-results}, CARF achieves the most favorable combination of include recall and F1 among all evaluated methods, correctly identifying 160 of the 166 eligible studies (recall 0.964) and missing six. The teacher-distilled reasoning model reaches the same recall, but only by forwarding nearly three times as many ineligible studies (325 vs.~119 false positives). CARF also has the highest F1 score (0.719), although its margin over label-only fine-tuning (0.714) is small. Compared with the zero-shot frontier models DeepSeek-V3.2, Grok-4.3, and Claude Sonnet-5, CARF matches or exceeds their recall and F1 while maintaining comparable precision.

These results also reveal a precision--recall trade-off among the remaining approaches. The label-only variant achieves slightly higher precision than CARF (0.588 vs.~0.573), and PubMedBERT achieves the highest precision overall (0.684) with the fewest false positives. These gains in precision, however, come with lower recall: label-only fine-tuning misses 15 eligible studies, and PubMedBERT misses 45. For systematic review screening this trade-off is consequential, because a false positive is removed at full-text review, whereas an eligible study excluded at the screening stage never reaches downstream reviewers.

However, not all false positives are equally costly. To examine them more closely, a post-hoc review was done of all the false positives, without knowing which model had included it, and each was rated as either clearly ineligible or ambiguous, that is, a borderline case a careful reviewer could reasonably send to full-text review. Roughly one in five of these articles was rated ambiguous, underscoring how context-dependent eligibility decisions are in this domain. Relative to label-only fine-tuning, CARF made a similar number of clearly ineligible inclusions but included more ambiguous studies, and relative to the teacher-distilled model, expert feedback removed clearly ineligible inclusions more readily than ambiguous ones. The full results of this review are reported in Appendix~\ref{app:fp-review}.

CARF consequently occupies a favorable operating point for first-pass systematic review screening, combining near-complete retrieval of eligible studies with a precision level that limits the additional burden on human reviewers. Its higher false-positive count relative to more conservative models represents a reasonable trade-off for the intended human-in-the-loop setting, particularly as part of that excess consists of borderline studies that are resolved through subsequent expert verification.

Beyond aggregate performance, analysis of the model's reasoning reveals a self-corrective behavior that appears after training on expert feedback. The adapted model sometimes initially considers evidence that could support inclusion, but then revisits that evidence against the screening criteria, identifies a contextual inconsistency or eligibility violation, and corrects its interpretation before producing the final decision. This behavior is strongly associated with correct exclusions: among the 834 excluded test articles, the model self-corrected on 271, of which only 3 resulted in a false-positive inclusion (1.1\%), compared with 116 of the 563 articles where it did not self-correct (20.6\%). The same behavior also accounts for 4 of the 6 eligible studies CARF missed, where the correction overturned an initially correct inclusion. Details and qualitative examples are given in Appendix~\ref{app:self-correction}.

\section{Discussion}

The results suggest that reasoning supervision offers a way to make language-model screening more compatible with the iterative and judgment-intensive nature of systematic reviews. Reasoning allows the model to connect evidence across an article and interpret that evidence in relation to the review criteria more thoroughly, improving the model performance when the evidence is ambiguous or conflicting. This creates an opportunity to shift part of the screening burden from exhaustive article-by-article assessment toward targeted human verification. By handling straightforward cases while making its reasoning available for inspection, a reasoning-based screener could accelerate the early stages of evidence synthesis without requiring experts to relinquish control over consequential eligibility decisions. More importantly, this efficiency does not require the model to make simpler or more aggressive screening decisions. By learning from how experts identify and resolve contextual errors, the model can handle nuanced eligibility criteria while still preserving the distinctions that matter for scientific screening. This makes reasoning-based screening a promising way to reduce manual screening effort without sacrificing the quality of evidence identification. 

We also found that not all forms of reasoning supervision were equally effective. Before developing the current expert-correction approach, we investigated a label-conditioned reasoning strategy in which the reasoning model was provided with the verified screening decision and asked to generate a rationale explaining that decision. This approach degraded the overall screening performance of the model. This finding is consistent with recent evidence that conditioning reasoning generation on a known answer can produce answer-conditioned rationalization, in which the model reasons backward from the target answer rather than deriving the answer from the available evidence \citep{lee2026answer}. This approach and its results are provided in Appendix~\ref{app:label-conditioned}. 

The amount of expert feedback also appears to influence adaptation performance. When CARF was retrained with only 100 or 500 of the 1,346 expert feedback notes, include recall rose from 0.922 with 100 notes to 0.964 with 500, while false positives showed no consistent trend (Appendix~\ref{app:feedback-amount}). Each configuration was trained once, so these differences are indicative rather than conclusive, but they suggest that broader exposure to expert corrections helps the model retain eligible studies.

CARF also holds the potential to address the growing gap between the volume of biomedical literature and the capacity of experts to synthesize it in a timely manner. By accelerating the initial identification of relevant studies, reasoning-based screening could shorten the time required to conduct and update systematic reviews, allowing researchers to incorporate emerging evidence more quickly. This is particularly important in rapidly evolving biomedical areas, where delays in evidence synthesis can slow the translation of new findings into clinical guidelines, research priorities, and public-health policies. By retaining expert oversight while reducing the amount of routine screening work, CARF offers a pathway toward more timely and scalable biomedical evidence synthesis.

\section{Limitations}

Several considerations provide important context for interpreting these findings. First, the expert screening decisions used as the reference standard may themselves contain errors. Prior work has shown that human reviewers are not perfectly consistent in systematic review screening, documenting non-trivial false inclusion and exclusion rates among expert reviewers during abstract screening \citep{wang2020errorrates}. Thus, the performance of CARF should be interpreted relative to the expert-annotated reference standard rather than as an absolute measure of correctness. This limitation is inherent to both human- and model-assisted screening evaluation and highlights the importance of continued expert oversight when applying automated approaches to evidence synthesis.

Second, CARF is evaluated within a single domain: systematic review screening for verbal autopsy and cause-of-death ascertainment. Although the framework is designed around a general mechanism for incorporating expert corrections into reasoning-model adaptation, the contextual distinctions and scientific criteria involved in VA/COD screening may differ substantially from those in other biomedical or specialized decision-making settings. Evaluation across additional systematic review domains and other expert-driven scientific tasks is therefore needed to determine whether the observed benefits generalize beyond this setting.

Third, our analysis of self-corrective reasoning demonstrates a strong association between self-correction and correct exclusion, but does not establish that self-correction itself causes improved screening performance. Self-correction may instead be a behavioral manifestation of other changes induced during model adaptation. Further controlled experiments would be needed to isolate the causal contribution of this behavior.

Fourth, the practical benefits of CARF must be considered alongside its computational and annotation costs. Although CARF improves screening performance on several metrics, zero-shot frontier models achieve comparable performance in some settings without requiring annotation of a subset by experts, expert-feedback collection or model fine-tuning. Depending on the relative importance of screening performance, cost efficiency, and implementation overhead, zero-shot models may therefore be a more suitable choice for certain applications.

Fifth, results obtained with large language models are difficult to reproduce exactly. The zero-shot frontier models were accessed through hosted APIs whose underlying versions may be updated or deprecated without notice, and even at temperature 0 their outputs are not guaranteed to be deterministic, so repeating our queries may not yield identical decisions. Our fine-tuned models use open-weight checkpoints and greedy decoding, which makes them more reproducible, but differences in hardware, inference software, and numerical precision can still produce small variations in generated reasoning and decisions. Reporting model versions and decoding parameters therefore reduces, but does not eliminate, this variability.

Finally, while our experiments indicate that increasing the amount of expert-corrected reasoning supervision may improve model performance, the present study does not establish how much expert feedback is sufficient or which characteristics of feedback are most important for effective adaptation. The quality, specificity, and consistency of expert corrections may influence what the model learns, and different experts may emphasize different aspects of the same screening decision. Our findings should therefore be viewed as an initial demonstration that expert-corrected reasoning can serve as a useful supervision signal for improving context-dependent systematic review screening, rather than as a definitive characterization of the amount or form of expert feedback required. Further work is needed to investigate these relationships. 

\section*{Ethical Considerations}

The use of large language models (LLMs) for systematic review screening raises concerns beyond the potential omission of relevant evidence, including overreliance on automated decisions, reduced human scrutiny, and the uncritical acceptance of model-generated judgments. Such reliance may allow errors and biases to propagate into evidence synthesis and, ultimately, influence scientific conclusions and policy decisions. CARF adopts a human-in-the-loop approach in which domain experts review model decisions and reasoning, provide corrections for disagreements, and guide model adaptation. This process keeps expert judgment central to screening and creates opportunities to identify and correct errors rather than treating model outputs as authoritative. In contrast, zero-shot approaches can be deployed without an explicit expert-feedback and correction stage, potentially encouraging greater reliance on automated decisions. However, human oversight is not infallible, as expert judgments can also be inconsistent or erroneous. Responsible deployment therefore requires meaningful expert involvement throughout the screening process, with model outputs treated as decision support rather than substitutes for scientific judgment and final human responsibility.

\section*{Acknowledgments}

Research reported in this publication was supported by the Eunice Kennedy Shriver National Institute of Child Health \& Human Development of the National Institutes of Health under Award Number R21HD119931. The content is solely the responsibility of the authors and does not necessarily represent the official views of the National Institutes of Health.
The authors also thank Lori Rosman, Information Specialist at the Johns Hopkins Welch Medical Library, for her assistance in developing the search strategy. 

\bibliography{custom}

\appendix

\section*{Appendix}
 
\section{Complete Search Strategy}
\label{app:search}
 
To identify studies on verbal autopsy, cause-of-death ascertainment, and artificial intelligence methods, we searched PubMed, Embase, Scopus, ACM Digital Library, and arXiv. The search strategy combined three core concepts: (1) verbal autopsy and cause-of-death terminology, (2) mortality-related terminology, and (3) artificial intelligence and machine learning methods. Database-specific syntax was adapted while preserving these concepts.
 
\subsection{PubMed Database}
 
\noindent\textit{Date run: 20Nov2025. Articles found: 3,739.}
 
\begin{lstlisting}
#1
("Autopsy"[Mesh:NoExp] OR "Death Certificates"[Mesh] OR "Cause of death"[mesh] OR "verbal autops*"[tiab] OR "death certificate*"[tiab] OR "autopsy report*"[tiab] OR "medical narrative*"[tiab] OR "cause of death"[tiab])
 
#2
("Mortality"[Mesh] OR "Death"[Mesh] OR "mortality"[tiab] OR "mortalities"[tiab] OR "fatal"[tiab] OR "fatality"[tiab] OR "fatalities"[tiab] OR "death"[tiab] OR "deaths"[tiab])
 
#3
("Artificial Intelligence"[Mesh] OR "Algorithms"[Mesh] OR "Computing Methodologies"[Mesh:NoExp] OR "Decision Making, Computer-Assisted"[Mesh] OR algorithm*[tiab] OR "artificial Intelligence"[tiab] OR "AI"[tiab] OR "comput* intelligen*"[tiab] OR "machine intelligence"[tiab] OR "computer reasoning"[tiab] OR "computer vision*"[tiab] OR "machine learning"[tiab] OR "deep learning"[tiab] OR "shallow learning"[tiab] OR "neural network*"[tiab] OR CNN[tiab] OR (generative[tiab] AND network*[tiab]) OR GAN[tiab] OR "computer-assisted"[tiab] OR "computer-aided"[tiab] OR "support vector machine*"[tiab] OR "support vector network*"[tiab] OR "computer methodologies"[tiab] OR "k Nearest Neighbor"[tiab] OR "K nearest neighbour"[tiab] OR kNN[tiab] OR "random forest*"[tiab] OR LLM[tiab] OR "Large language model*"[tiab] OR "natural language process*"[tiab] OR BERT[tiab] OR "ClinicalBERT"[tiab] OR GPT[tiab])
 
#4
#1 AND #2 AND #3
\end{lstlisting}
 
\subsection{Embase Database}
 
\noindent\textit{Date run: 20Nov2025. Articles found: 7,662.}
 
\begin{lstlisting}
#1
'autopsy'/de OR 'verbal autopsy'/exp OR 'death certificate'/exp OR 'cause of death'/exp OR ("verbal autops*" OR "death certificate*" OR "autopsy report*" OR "medical narrative*" OR "cause of death"):ti,ab,kw
 
#2
'mortality'/exp OR 'death'/exp OR ("mortality" OR "Mortalities" OR "fatal" OR "fatality" OR "fatalities" OR "death" OR "deaths"):ab,ti,kw
 
#3
'artificial intelligence'/exp OR 'algorithm'/exp OR 'computer analysis'/exp OR 'decision support system'/exp OR (algorithm* OR "artificial Intelligence" OR "AI" OR "comput* intelligen*" OR "machine intelligence" OR "computer reasoning" OR "computer vision*" OR "machine learning" OR "deep learning" OR "shallow learning" OR "neural network*" OR CNN OR (generative NEXT/2 network*) OR GAN OR "computer-assisted" OR "computer-aided" OR "support vector machine*" OR "support vector network*" OR "computer methodologies" OR "k Nearest Neighbor" OR "K nearest neighbour" OR kNN OR "random forest*" OR LLM OR "Large language model*" OR "natural language process*" OR BERT OR "ClinicalBERT" OR GPT):ab,ti,kw
 
#4
#1 AND #2 AND #3
\end{lstlisting}
 
\subsection{Scopus Database}
 
\noindent\textit{Date run: 20Nov2025. Articles found: 5,368.}
 
\begin{lstlisting}
(TITLE-ABS-KEY ( ( "verbal autops*" OR "death certificate*" OR "autopsy report*" OR "medical narrative*" OR "cause of death" ) )) AND (TITLE-ABS-KEY ( ( "mortality" OR "mortalities" OR "fatal" OR "fatality" OR "fatalities" OR "death" OR "deaths" ) )) AND (TITLE-ABS-KEY ( ( algorithm* OR "artificial Intelligence" OR "AI" OR "comput* intelligen*" OR "machine intelligence" OR "computer reasoning" OR "computer vision*" OR "machine learning" OR "deep learning" OR "shallow learning" OR "neural network*" OR CNN OR ( generative PRE/2 network* ) OR GAN OR "computer-assisted" OR "computer-aided" OR "support vector machine*" OR "support vector network*" OR "computer methodologies" OR "k Nearest Neighbor" OR "K nearest neighbour" OR kNN OR "random forest*" OR LLM OR "Large language model*" OR "natural language process*" OR BERT OR "ClinicalBERT" OR GPT ) )) AND NOT (index (medline OR embase))
\end{lstlisting}
 
\subsection{ACM Digital Library Database}
 
\noindent\textit{Date run: 20Nov2025. Articles found: 877 records.}
 
\begin{lstlisting}
("verbal autopsy" OR "verbal autopsies" OR "death certificate" OR "death certificates" OR "autopsy report" OR "autopsy reports" OR "medical narrative" OR "medical narratives" OR "cause of death") AND ("mortality" OR "mortalities" OR "fatal" OR "fatality" OR "fatalities" OR "death" OR "deaths")
\end{lstlisting}
 
\subsection{ArXiv Database}
 
\noindent\textit{Date run: 3Dec2025. Articles found: 1,007 results.}
 
\begin{lstlisting}
# write the search strategy as query
query9 <-   '"verbal autopsy" OR "verbal autopsies" OR "death certificate" OR "death certificates" OR "autopsy report" OR "autopsy reports"
            OR (("medical narrative" OR "medical narratives") AND (mortalit* OR fatal* OR death*)) OR "cause of death" AND (mortalit* OR fatal*) OR
            ("cause of death" AND (algorithm* OR artificial OR Intelligence OR Intelligent OR "AI" OR computer OR "machine learning"
             OR  "deep learning" OR "shallow learning" OR "neural network" OR "neural networks" OR CNN OR "generative network" OR "generative networks"
             OR GAN OR "support vector" OR "support vectors" OR "k Nearest Neighbor" OR "k Nearest Neighbors" OR "K nearest neighbour" OR "K nearest neighbours"
             OR kNN OR "random forest" OR "random forests" OR LLM OR "Large language" OR "natural language" OR BERT OR "ClinicalBERT" OR GPT))'
 
# Run the query
results_df <- arxiv_search(query = query9, limit = 1200, batchsize = 50) #Total 1007 results with this
 
# Extract year from updated date
results_df$year <- substr(results_df$updated, 1, 4)
 
# Write RIS file to a chosen location
ris_file_path <- "path/to/arxiv_results_upd.ris"
 
write_refs(results_df, file = ris_file_path, format = "ris")
\end{lstlisting}

\section{Title and Abstract Screening Criteria}
\label{app:screening}

This systematic review examines the use of artificial intelligence (AI) and machine learning methods for cause-of-death ascertainment, particularly in comparison with traditional approaches such as verbal autopsy (VA) and vital registration (VR). The project aims to systematically identify and assess studies reporting or assessing cause-of-death data and, at the full-text stage, characterize studies that use AI-based approaches for cause-of-death ascertainment. The screening protocol was therefore designed to identify relevant mortality studies for subsequent full-text assessment.

Title and abstract screening was performed using predefined exclusion criteria. Each record was assigned one mutually exclusive screening decision. Records meeting any exclusion criterion were excluded at this stage; records for which none of the exclusion criteria applied proceeded to full-text screening.

\begin{center}
\small
\begin{tabular}{p{0.30\columnwidth}p{0.58\columnwidth}}
\hline
\textbf{Screening decision} & \textbf{Description} \\
\hline
Include & The study did not meet any of the predefined exclusion criteria and proceeded to full-text screening. \\
Missing abstract or full text & The publication lacked an abstract or accessible full text required for screening. \\
Non-human study & The study was conducted on non-human subjects, such as animals or bacteria. \\
No cause-of-death data & The study did not present or ascertain cause-of-death data as absolute numbers, rates, risks, or ratios. \\
\textless{}15 deaths with known cause & The study reported fewer than 15 deaths with cause-of-death information. \\
Duplicate & The record had already been identified in the database and was excluded as a duplicate. \\
\hline
\end{tabular}
\end{center}

\subsection*{Eligibility for Full-Text Screening}

Studies proceeded to full-text screening if none of the title and abstract exclusion criteria applied. At this stage, additional study characteristics were recorded, including study language, whether the study population was representative of the general population or a subpopulation, whether children under five years were included, the number of causes of death reported, whether an AI cause-ascertaining algorithm was used, and whether performance comparisons between AI and traditional VA/VR methods were reported.

\section{Implementation Details}
\label{app:impl}
\setcounter{table}{0}
\renewcommand{\thetable}{C\arabic{table}}
 
\subsection{Experimental Setup}

All experiments were conducted on a high-performance computing cluster. Each training or evaluation job used a single GPU; independent jobs were sometimes run concurrently on separate GPUs.

\paragraph{Hardware.} All reported experiments were run on NVIDIA H200 GPUs (141\,GB), one GPU per job. Reasoning trace generation with the teacher model used the vLLM inference engine; student model fine-tuning, evaluation, and the PubMedBERT baseline used the Hugging Face stack described below.

\paragraph{Computational budget.} Fine-tuning one student model for three epochs took approximately 1.6 GPU-hours, and evaluating one model on the 1,000-article test set took approximately 2.1 GPU-hours. Across all experiments, including the fine-tuned variants reported in the appendices and preliminary and exploratory runs, the total computational budget was approximately 80 GPU-hours.

\paragraph{Software.} Experiments were implemented in Python 3.11 with PyTorch 2.9.0 \citep{paszke2019pytorch} and CUDA 12.8. Teacher inference used vLLM 0.11.2 \citep{kwon2023vllm}, while student model fine-tuning was implemented using Hugging Face Transformers 4.57.6 \citep{wolf2020transformers} together with PEFT 0.19.1 \citep{mangrulkar2022peft}. The Hugging Face Trainer API was used for optimization and checkpoint management. Fine-tuning of the reasoning models used Liger Kernel 0.8.1's fused linear cross-entropy \citep{hsu2024liger} to reduce GPU memory use; this changes how the loss is computed in memory but not its value. At evaluation, student models generated their outputs with greedy decoding (no sampling), with a limit of 3,072 new tokens for the reasoning models and 64 for the label-only model, whose output is a single decision line. The PubMedBERT baseline \citep{gu2022} was implemented using AutoModelForSequenceClassification. All pretrained model checkpoints were stored locally on the cluster filesystem to eliminate runtime model downloads.

\paragraph{Evaluation and statistical analysis.} All metrics are computed for the Include class with our own scoring script, which treats every exclusion category as a negative. Recall, specificity, precision, and F1 follow their standard definitions. Confidence intervals for recall and for the proportions in Appendix~\ref{app:fp-review} are exact 95\% Clopper--Pearson intervals \citep{clopper1934binomial}, computed with SciPy 1.18.1 \citep{virtanen2020scipy}. The comparison of removal rates in Appendix~\ref{app:fp-review} uses a two-sided Fisher's exact test \citep{fisher1922interpretation}; both tests were computed directly from its exact hypergeometric distribution.

\subsection{Training Hyperparameters}

All fine-tuned language models used the values below; no hyperparameter search was performed, and neither the validation nor the test set was used to choose them. Table~\ref{tab:hparams} lists the LoRA settings shared by the three fine-tuned language models (CARF, label-only, and teacher-distilled reasoning). Hyperparameters for the PubMedBERT baseline are reported separately with the baseline details in Appendix~\ref{app:baselines} (Table~\ref{tab:hparams-bert}).

\begin{center}
\small
\captionof{table}{LoRA fine-tuning hyperparameters for the student language models (CARF, label-only, and teacher-distilled reasoning).}
\label{tab:hparams}
\begin{tabular}{p{0.38\columnwidth}p{0.5\columnwidth}}
\hline
\textbf{Hyperparameter} & \textbf{Value} \\
\hline
Base model & DeepSeek-R1-Distill-Qwen-14B \\
Fine-tuning method & LoRA \\
LoRA rank (r) & 16 \\
LoRA alpha & 32 \\
LoRA dropout & 0.05 \\
Target modules & All attention and feed-forward projection layers \\
Attention projections & {\footnotesize\texttt{q\_proj, k\_proj, v\_proj, o\_proj}} \\
Feed-forward projections & {\footnotesize\texttt{gate\_proj, up\_proj, down\_proj}} \\
Trainable parameters & 68.8M (0.46\% of 14.8B) \\
Precision & bfloat16 \\
Gradient checkpointing & Enabled \\
Training epochs & 3 \\
Per-device batch size & 1 \\
Gradient accumulation steps & 16 \\
Effective batch size & 16 \\
Learning rate & $1 \times 10^{-4}$ \\
Learning rate scheduler & Cosine decay \\
Warm-up ratio & 5\% \\
Weight decay & 0 \\
Max gradient norm & 1.0 \\
Maximum sequence length & 8,192 tokens \\
Random seed & 42 \\
Training objective & Completion-only supervised fine-tuning \\
Checkpoint selection & Lowest validation loss, evaluated each epoch \newline ({\scriptsize\texttt{load\_best\_model\_at\_end}}) \\
\hline
\end{tabular}
\end{center}

\section{Screening Instructions}
\label{app:prompts}

\subsection{Model-Assisted Verification Prompt}
\label{app:prompt-verification}

The exact prompt used during the model-assisted annotation verification stage is reproduced verbatim below.

\begin{lstlisting}[
  basicstyle=\ttfamily\footnotesize,
  breaklines=true,
  frame=single,
  framesep=6pt,
  xleftmargin=6pt,
  xrightmargin=6pt,
  columns=fullflexible,
  keepspaces=true,
  caption={Model-assisted verification screening prompt.},
  label={lst:verification-prompt}
]
You are an expert systematic reviewer working on a scientific project
systematically extracting information from peer-reviewed and grey
literature and pre-prints to identify artificial intelligence
algorithms used to ascertain causes of death from various data sources.

ABOUT CAUSE OF DEATH (COD) DATA SOURCES:
Cause of death can be determined through several different mechanisms,
each suited to different resource settings:
1. Medically Certified Cause of Death: In high-income and
   well-resourced countries with strong Vital Registration (VR) systems
   (also called Civil Registration and Vital Statistics (CRVS)
   systems), deaths are typically certified by a physician or physician
   in training at the time of death and recorded through civil
   registration. This is generally considered the most reliable source
   of COD data.
2. Verbal Autopsy (VA): In settings where VR systems are weak or
   absent, commonly in low- and middle-income countries (LMICs), VA is
   used instead, relying on structured and sometimes open-ended
   narrative interviews with family members to infer a probable COD.
3. Other clinical/administrative records: clinical charts or their
   extraction, electronic medical records, prescriptions,
   socio-economic, behavioral, or other data from the deceased
   individuals.

This systematic review considers studies presenting COD data and/or
AI/ML-based ascertainment methods from ANY of these sources and
settings - it is NOT exclusively VA or LMIC-based, nor exclusively
high-income. A study conducted in a high-income country using medically
certified COD data is just as relevant as a study using VA in an LMIC,
provided it meets the inclusion criteria. Do NOT exclude a study merely
because it does not use verbal autopsy or is not set in an LMIC.

PURPOSE:
As part of this landscape analysis, this systematic review
identifies all studies that present COD data and/or report AI/ML
algorithms for cause ascertainment, to map how AI-based ascertainment
methods have been used across different data sources.

YOUR TASK:
You will be given ONLY the title and abstract of a study - with NO
prior screening decision. Screen it yourself. Read the title and
abstract carefully and decide, from the abstract alone, whether it
should be INCLUDED for full-text review or EXCLUDED, and if excluded,
for which specific reason.

Apply these EXCLUSION CRITERIA in order. Exclude at the FIRST criterion
the study meets:
1. NON-HUMAN STUDY: the study does not involve human subjects
   -> EXCLUDE_NON_HUMAN
2. LESS THAN 15 DEATHS WITH KNOWN CAUSE: the study reports fewer than
   15 deaths with known cause of death -> EXCLUDE_UNDER_15_DEATHS
3. MISSING ABSTRACT OR FULL-TEXT: the abstract or full text is
   unavailable (e.g. the abstract is empty or a placeholder such as
   "[No abstract available]") -> EXCLUDE_MISSING_TEXT
4. NO CAUSE-OF-DEATH DATA: This review is about ASCERTAINING CAUSES OF
   DEATH - i.e. determining what people died FROM. To be INCLUDED, a
   study must present or ascertain cause-of-death information: data
   about the CAUSES of death (what people died of), whether as specific
   causes, cause-specific counts/rates/risks/ratios, or the study's aim
   is to assign/ascertain/classify cause of death from some data source.

CRITICAL DISTINCTION - mortality as an OUTCOME is NOT cause-of-death
data. EXCLUDE as EXCLUDE_NO_COD if the study merely:
  - PREDICTS mortality/death RISK as a machine-learning outcome (e.g.
    reports AUC/C-statistic for predicting whether a patient dies), or
  - reports ALL-CAUSE mortality as an endpoint/outcome WITHOUT
    identifying the causes of death (e.g. "intervention reduced
    all-cause mortality by 14%"), or
  - uses death as a prediction target, survival endpoint, or risk score
    rather than ascertaining what people died OF, or
  - reports only disease prevalence, incidence, risk factors,
    diagnostic accuracy, or clinical outcomes.

The test is NOT "does the abstract mention death or mortality" - many
excluded studies mention mortality. The test is: does the study present
or ascertain the CAUSES of death (what people died from)? A
stroke-mortality-prediction model, an all-cause-mortality risk score,
or a study reporting that a treatment reduced mortality does NOT
ascertain causes of death -> EXCLUDE_NO_COD. A study that reports
deaths broken down by cause, or whose purpose is to determine/classify
cause of death, DOES qualify -> passes this criterion.

When unsure whether the study ascertains CAUSES of death (versus merely
predicting or reporting mortality), lean toward EXCLUDE_NO_COD - do not
include on the basis of mortality being mentioned.

Mark INCLUDE if the study satisfies ALL requirements: human subjects,
abstract available and readable, >=15 deaths with known cause, AND the
study presents or ascertains CAUSES of death (what people died from) -
not merely predicts or reports mortality as an outcome. If the study
does not ascertain causes of death, the correct decision is
EXCLUDE_NO_COD.

Reason step by step BEFORE giving your decision:
1. Briefly describe what the paper is about (population, setting,
   methodology).
2. Work through each exclusion criterion explicitly, citing specific
   evidence from the title or abstract.
3. State your final decision clearly.
Ground every claim in specific wording from the abstract. Use only what
the abstract states - do not use outside knowledge about the paper.

Return your answer as a JSON object with EXACTLY these keys:
{
  "reasoning": "<your step-by-step screening reasoning, 150-300 words, citing the abstract>",
  "decision": "<one of: INCLUDE | EXCLUDE_NON_HUMAN | EXCLUDE_UNDER_15_DEATHS | EXCLUDE_MISSING_TEXT | EXCLUDE_NO_COD | INSUFFICIENT_INFO>",
  "confidence": "<high | medium | low>",
  "key_evidence": "<the specific phrase(s) from the abstract that drove the decision>"
}
Return ONLY the JSON object, nothing before or after it.
\end{lstlisting}

\subsection{Blind Screening and Reasoning Generation}
\label{app:prompt-blind}

The exact prompt supplied to the model for blind screening and reasoning generation is given below.

\begin{lstlisting}[
  basicstyle=\ttfamily\footnotesize,
  breaklines=true,
  frame=single,
  framesep=6pt,
  xleftmargin=6pt,
  xrightmargin=6pt,
  columns=fullflexible,
  keepspaces=true,
  caption={Blind screening and reasoning-generation prompt.},
  label={lst:blind-prompt}
]
You are an expert systematic reviewer working on a project that
systematically extracts information from peer-reviewed and grey
literature and pre-prints to identify artificial intelligence
algorithms used to ascertain causes of death from various data sources.

ABOUT CAUSE OF DEATH (COD) DATA SOURCES:
Cause of death can be determined through several mechanisms:
1. Medically Certified Cause of Death: in high-income / well-resourced
   settings with strong Vital Registration (VR) or Civil Registration
   and Vital Statistics (CRVS) systems, deaths are certified by a
   physician and recorded through civil registration. Generally the
   most reliable source.
2. Verbal Autopsy (VA): where VR is weak or absent (often LMICs), a
   probable COD is inferred from structured interviews with family
   members.
3. Other clinical/administrative records: clinical charts, electronic
   medical records, prescriptions, or socio-economic/behavioral data
   from the deceased.

This review considers studies presenting COD data and/or AI/ML-based
ascertainment methods from ANY of these sources and settings - it is
NOT restricted to VA-only, LMIC-only, or high-income-only studies. A
high-income study using medically certified COD data is as relevant as
an LMIC study using VA, provided it meets the inclusion criteria.

PURPOSE:
Identify all studies that present COD data and/or report AI/ML
algorithms for cause ascertainment, to map how AI-based methods have
been used to ascertain COD across data sources.

YOUR TASK:
Read the title and abstract below and DECIDE whether this paper should
continue to full-text screening. Reason step by step through the
exclusion criteria, citing specific evidence from the title or
abstract, then state your decision.

EXCLUSION CRITERIA (checked in order - EXCLUDE if ANY is met):
1. Non-human study: the study does not involve human subjects.
2. Fewer than 15 deaths with known cause: the study reports fewer than
   15 deaths with known COD.
3. No cause of death data: COD is not presented as absolute numbers,
   rates, risks, or ratios. Studies that report only survival outcomes,
   clinical outcomes, risk PREDICTION, or disease prevalence - without
   breaking down what people actually died FROM - are excluded. (A
   study that predicts who will die, or models mortality risk, does NOT
   by itself present cause-of-death data.)

If none of the exclusion criteria are met, INCLUDE the paper.

---
Title: {title}

Abstract: {abstract}
---

Think through the criteria, then end your response with EXACTLY one line
in this format:

DECISION: <one of INCLUDE | EXCLUDE_NON_HUMAN | EXCLUDE_UNDER_15 | EXCLUDE_NO_COD>
\end{lstlisting}

\subsection{Training Prompt for All Models During Fine-Tuning and Evaluation}
\label{app:prompt-training}

The exact prompt used for all models during fine-tuning and evaluation is reproduced verbatim below.

\begin{lstlisting}[
  basicstyle=\ttfamily\footnotesize,
  breaklines=true,
  frame=single,
  framesep=6pt,
  xleftmargin=6pt,
  xrightmargin=6pt,
  columns=fullflexible,
  keepspaces=true,
  caption={Training and evaluation prompt provided to all models.},
  label={lst:training-prompt}
]
You are an expert systematic reviewer working on a scientific project
systematically extracting information from peer-reviewed and grey
literature and pre-prints to identify artificial intelligence
algorithms used to ascertain causes of death from various data sources.

ABOUT CAUSE OF DEATH (COD) DATA SOURCES:
Cause of death can be determined through several different mechanisms,
each suited to different resource settings:
1. Medically Certified Cause of Death: In high-income and
   well-resourced countries with strong Vital Registration (VR) systems
   (also called Civil Registration and Vital Statistics (CRVS)
   systems), deaths are typically certified by a physician or physician
   in training at the time of death and recorded through civil
   registration. This is generally considered the most reliable source
   of COD data.
2. Verbal Autopsy (VA): In settings where VR systems are weak or
   absent, commonly in low- and middle-income countries (LMICs), VA is
   used instead, relying on structured and sometimes open-ended
   narrative interviews with family members to infer a probable COD.
3. Other clinical/administrative records: clinical charts or their
   extraction, electronic medical records or other data from the
   deceased individuals.

This systematic review considers studies presenting COD data and/or
AI/ML-based ascertainment methods from ANY of these sources and
settings - it is NOT exclusively VA or LMIC-based, nor exclusively
high-income. A study conducted in a high-income country using medically
certified COD data is just as relevant as a study using VA in an LMIC,
provided it meets the inclusion criteria. Do NOT exclude a study merely
because it does not use verbal autopsy or is not set in an LMIC.

PURPOSE:
As part of this project's landscape analysis, this systematic review
identifies all studies that present COD data and/or report AI/ML
algorithms for cause ascertainment, to map how AI-based ascertainment
methods have been used across different data sources.

YOUR TASK:
You will be given ONLY the title and abstract of a study - with NO
prior screening decision. Screen it yourself. Read the title and
abstract carefully and decide, from the abstract alone, whether it
should be INCLUDED for full-text review or EXCLUDED, and if excluded,
for which specific reason.

EXCLUSION CRITERIA:
Apply these in order. Exclude at the FIRST criterion the study meets:
1. NON-HUMAN STUDY: the study does not involve human subjects
   -> EXCLUDE_NON_HUMAN
2. LESS THAN 15 DEATHS WITH KNOWN CAUSE: the study reports fewer than
   15 deaths with known cause of death -> EXCLUDE_UNDER_15_DEATHS
3. MISSING ABSTRACT OR FULL-TEXT: the abstract or full text is
   unavailable (e.g. the abstract is empty or a placeholder such as
   "[No abstract available]") -> EXCLUDE_MISSING_TEXT
4. NO CAUSE-OF-DEATH DATA: This review is about ASCERTAINING CAUSES OF
   DEATH - i.e. determining what people died FROM. To be INCLUDED, a
   study must present or ascertain cause-of-death information: data
   about the CAUSES of death (what people died of), whether as specific
   causes, cause-specific counts/rates/risks/ratios, or the study's aim
   is to assign/ascertain/classify cause of death from some data source.

CRITICAL DISTINCTION - mortality as an OUTCOME is NOT cause-of-death
data. Exclude as EXCLUDE_NO_COD if the study merely:
  - PREDICTS mortality/death RISK as a machine-learning outcome (e.g.
    reports AUC/C-statistic for predicting whether a patient dies), or
  - reports ALL-CAUSE mortality as an endpoint/outcome WITHOUT
    identifying the causes of death (e.g. "intervention reduced
    all-cause mortality by 14%"), or
  - uses death as a prediction target, survival endpoint, or risk score
    rather than ascertaining what people died OF, or
  - reports only disease prevalence, incidence, risk factors,
    diagnostic accuracy, or clinical outcomes.

The test is NOT "does the abstract mention death or mortality" - many
excluded studies mention mortality. The test is: does the study present
or ascertain the CAUSES of death (what people died from)? A
stroke-mortality-prediction model, an all-cause-mortality risk score,
or a study reporting that a treatment reduced mortality does NOT
ascertain causes of death -> EXCLUDE_NO_COD. A study that reports
deaths broken down by cause, or whose purpose is to determine/classify
cause of death, DOES qualify -> passes this criterion.

When unsure whether the study ascertains CAUSES of death (versus merely
predicting or reporting mortality), lean toward EXCLUDE_NO_COD - do not
include on the basis of mortality being mentioned.

Mark INCLUDE if the study satisfies ALL requirements: human subjects,
abstract available and readable, >=15 deaths with known cause, AND the
study presents or ascertains CAUSES of death (what people died from) -
not merely predicts or reports mortality as an outcome. If the study
does not ascertain causes of death, the correct decision is
EXCLUDE_NO_COD.

Ground every claim in specific wording from the abstract. Use only what
the abstract states - do not use outside knowledge about the paper.
\end{lstlisting}

\smallskip
\noindent\textit{Note.} This prompt's purpose is consistent with our evaluation focus on maximizing includes and recall for the Include class, ensuring that all relevant articles are retained for full-text screening. Although the prompt instructs the model to ``lean toward \texttt{EXCLUDE\_NO\_COD} when uncertain,'' this applies only to cases where mortality or cause of death appears as background or contextual framing rather than as the study's subject. For example, an article that notes ``heart disease is a leading cause of death...'' but does not actually ascertain or report how cause of death was determined for deceased patients. This is consistent with the human annotation policy used throughout this review.

\section{Baseline Implementation Details}
\label{app:baselines}
\renewcommand{\thetable}{\thesection\arabic{table}}
\setcounter{table}{0}

\subsection{PubMedBERT Classification Baseline}

To compare CARF against a conventional document classification approach, we fine-tune PubMedBERT \citep{gu2022} on the same training, validation, and test splits. We train a binary classifier predicting Include versus Exclude. The model is fine-tuned end-to-end using standard hyperparameters for transformer-based biomedical classifiers, with all model settings fixed across experiments. The full configuration is given in Table~\ref{tab:hparams-bert}.

\begin{center}
\small
\captionof{table}{Fine-tuning hyperparameters for the PubMedBERT baseline. Values marked $^\dagger$ are the Hugging Face Trainer defaults, which were not changed.}
\label{tab:hparams-bert}
\begin{tabular}{p{0.38\columnwidth}p{0.5\columnwidth}}
\hline
\textbf{Hyperparameter} & \textbf{Value} \\
\hline
Base model & BiomedBERT (base, uncased; PubMed abstracts and PMC full text) \\
Parameters & $\approx$110M (all fine-tuned) \\
Task head & Sequence classification \\
Output classes & 2 (Include, Exclude) \\
Input & ``\emph{title}. \emph{abstract}'' \\
Maximum sequence length & 512 tokens (longer inputs truncated) \\
Training data & Same training articles as the label-only model \\
Precision & bfloat16 \\
Training epochs & 4 \\
Batch size & 16 \\
Learning rate & $2 \times 10^{-5}$ \\
Optimizer & AdamW$^\dagger$ \\
Learning rate scheduler & Linear decay$^\dagger$ \\
Warm-up & None$^\dagger$ \\
Weight decay & 0.01 \\
Max gradient norm & 1.0$^\dagger$ \\
Random seed & 42 \\
Training objective & Cross-entropy classification \\
Checkpoint selection & Lowest validation loss, evaluated each epoch \newline ({\scriptsize\texttt{load\_best\_model\_at\_end}}) \\
\hline
\end{tabular}
\end{center}

\subsection{Label-only Fine-tuning Baseline}

We fine-tune DeepSeek-R1-Distill-Qwen-14B using conventional label supervision. The model is trained on the same 3,861 training articles used for our reasoning model (3,858 after 3 rows were dropped in preprocessing) and employs the same prompt template, LoRA configuration, optimization hyperparameters, and sequence length budget (Table~\ref{tab:hparams}). The only difference is the training target where the model is trained to predict the final screening decision (i.e., Include or the corresponding exclusion category). No reasoning traces or expert feedback are provided during fine-tuning.

\subsection{Reasoning-only Fine-tuning (teacher-distilled) Baseline}

We fine-tune DeepSeek-R1-Distill-Qwen-14B on the teacher model's reasoning (DeepSeek-R1-Distill-Qwen-32B), with no human supervision. Here we use the traces generated by the teacher model at the blind screening stage (Section~\ref{sec:blind-screening}). This model employs the same prompt template, LoRA configuration, optimization hyperparameters, and sequence length budget as our reasoning model (Table~\ref{tab:hparams}), and the same target serialization: the reasoning trace followed by the final decision. The only difference from CARF is the supervision signal where the training target uses the teacher model's blind decision rather than the verified reviewer label, and no expert feedback is provided. This baseline therefore captures reasoning-format fine-tuning in the complete absence of expert input, isolating the contribution of expert reasoning supervision.

\subsection{Zero-shot Language Model Baseline}

To compare CARF against strong inference-time reasoning approaches, we additionally evaluate 3 language models in a zero-shot setting. Each model receives the same screening instructions, review protocol, article title, and abstract as input, and is asked to produce a screening decision without task-specific fine-tuning. The models evaluated are DeepSeek-V3.2, Grok-4.3, and Claude Sonnet-5. Parameter counts are not disclosed for these hosted models. All models were queried through a LiteLLM proxy with no local GPU usage, within a total budget of 3,000 API calls (3 models $\times$ 1 prompt $\times$ 1{,}000 records). No hyperparameter search was performed. We requested temperature 0 and set the maximum number of output tokens to 600 (900 for Claude Sonnet-5), leaving all other sampling parameters at their provider defaults.

\section{Label-Conditioned Reasoning Approach}
\label{app:label-conditioned}
\renewcommand{\thetable}{\thesection\arabic{table}}
\setcounter{table}{0}

\subsection{Methodology}
\label{app:lc-eval}
Using the verified screening decisions in the training set ($n=3{,}861$), we construct an expert reasoning supervision dataset using a student-teacher framework \citep{NEURIPS2023_c6afe9a5}. DeepSeek-R1-Distill-Qwen-32B model is used to generate rationales on decisions for all 4,161 training and validation articles. Reasoning generation is label-conditioned, aiming to encourage the model to reconstruct the rationale underlying the verified expert decision while screening an article.

Then, the expert review is conducted on a stratified sample of 200 reasoning traces from the training set, comprising 146 articles excluded because no cause-of-death data were presented, 44 included articles, 6 studies with fewer than 15 deaths with known cause, 3 non-human studies, and 1 duplicate record. The sample is evaluated using a 2+1 review protocol, in which two primary reviewers independently assess each model-generated rationale and a third reviewer adjudicates disagreements. Reviewer feedback identifying incomplete, incorrect, or unsupported reasoning is paired with the corresponding teacher-generated rationale to construct the expert reasoning supervision dataset used for student model adaptation. 66 articles from this subset contained expert feedback. This is the only feedback provided for fine-tuning and complements the reasoning by capturing the evidence, eligibility criteria, and contextual considerations that influenced the expert's decision.

The resulting supervision consists of tuples containing the article title, abstract, verified screening decision, teacher-generated reasoning, and expert feedback for those 66 articles. This expert-augmented reasoning dataset forms the supervision used for reasoning model adaptation. This dataset contains 1.7\% expert feedback/corrections.

We adapt DeepSeek-R1-Distill-Qwen-14B as the student model using Low-Rank Adaptation (LoRA). Each training instance consists of the systematic review screening instructions, article title, and abstract as input, with the corresponding expert-augmented reasoning trace (if it exists) and screening decision as the target output. During training, loss is computed only over the target sequence, ensuring that supervision is derived exclusively from the expert reasoning trace and screening decision. LoRA adapters are applied to both the attention and feed-forward projection layers while the pretrained model weights remain frozen. Because CARF requires learning both context-dependent evidence associations and expert reasoning behavior, we adapt both sets of projection layers during fine-tuning.

\smallskip
\noindent\textit{Note:} This approach uses the same training, validation, and test splits, as well as the same hyperparameters, prompts, and training and screening instructions used to develop and evaluate the blind screening methodology presented in the paper.

\subsection{Evaluation}

Because systematic reviews prioritize identifying all relevant studies, recall (sensitivity) for the Include class serves as the primary evaluation metric: false negatives correspond to eligible studies incorrectly excluded, which may compromise the completeness of the review and are typically unrecoverable in later screening stages. False positives, by contrast, are corrected at low cost during full-text review.

\subsection{Results}

\begin{table*}[t]
\centering
\small
\setlength{\tabcolsep}{6pt}
\begin{tabular}{lcccccc}
\toprule
\textbf{Method} & \textbf{Rec.} & \textbf{Spec.} & \textbf{Prec.} & \textbf{F1} & \textbf{FP} & \textbf{Missed} \\
\midrule
\multicolumn{7}{l}{\textit{Fine-tuned (ours, 14B)}} \\
Label Conditioned Reasoning       & 0.958 & 0.787 & 0.472 & 0.632 & 178 & 7/166 \\
Label-only                        & 0.910 & 0.873 & 0.588 & 0.714 & 106 & 15/166 \\
\midrule
\multicolumn{7}{l}{\textit{Zero-shot frontier}} \\
Claude Sonnet-5                   & 0.922 & 0.860 & 0.567 & 0.702 & 117 & 13/166 \\
DeepSeek-V3.2                     & 0.940 & 0.821 & 0.511 & 0.662 & 149 & 10/166 \\
Grok-4.3                          & 0.922 & 0.853 & 0.554 & 0.692 & 123 & 13/166 \\
\bottomrule
\end{tabular}
\caption{Include-class screening performance of the label-conditioned reasoning variant compared with the label-only fine-tuned model and zero-shot frontier models on the held-out test set.}
\label{tab:label-conditioned-results}
\end{table*}

The comparison of the fine-tuned reasoning variants shows how the source of reasoning supervision affects screening behavior. CARF and the label-conditioned model reach nearly the same recall (0.964 vs.\ 0.958; 160 vs.\ 159 of 166 eligible studies), but label-conditioned reasoning produces 178 false positives compared with 119 for CARF, lowering precision from 0.573 to 0.472 and F1 from 0.719 to 0.632. Providing the target screening label before generating reasoning therefore does not yield more reliable screening, despite comparable recall. The teacher-distilled model, trained on blind reasoning traces without expert correction, matches CARF's recall but produces 325 false positives (precision 0.330, F1 0.492), the lowest precision and F1 among all evaluated models. Label-conditioned reasoning thus falls between the two: conditioning on the label reduces false positives relative to uncorrected blind reasoning, but considerably less than expert correction does. These results support blind reasoning with expert correction as the more reliable approach for systematic-review screening, particularly when high recall is needed while keeping the burden of subsequent human review manageable.

\section{Blinded Expert Review of False Positives}
\label{app:fp-review}
\setcounter{table}{0}
\renewcommand{\thetable}{\thesection\arabic{table}}
\setcounter{figure}{0}
\renewcommand{\thefigure}{\thesection\arabic{figure}}

\subsection{Procedure}

We pooled every test article that at least one evaluated model included but that the reference label marks as excluded, across all seven models in Table~\ref{tab:screening-results}. This yielded 347 unique articles. Each article was presented once, in random order, showing only its title and abstract; the reviewer could not see which model or models had included it. A domain expert rated each article as either \emph{clearly ineligible} (the abstract presents no cause-of-death data and does not aim to ascertain cause of death, or a different exclusion criterion clearly applies) or \emph{ambiguous} (a careful reviewer could reasonably send it to full-text review). Both categories were defined before the review began. Ratings were then linked back to the models that had included each article.

Of the 347 articles, 77 (22.2\%) were rated ambiguous and 270 clearly ineligible.

\subsection{False Positives by Model}

\begin{figure}[t]
  \centering
  \includegraphics[width=\columnwidth]{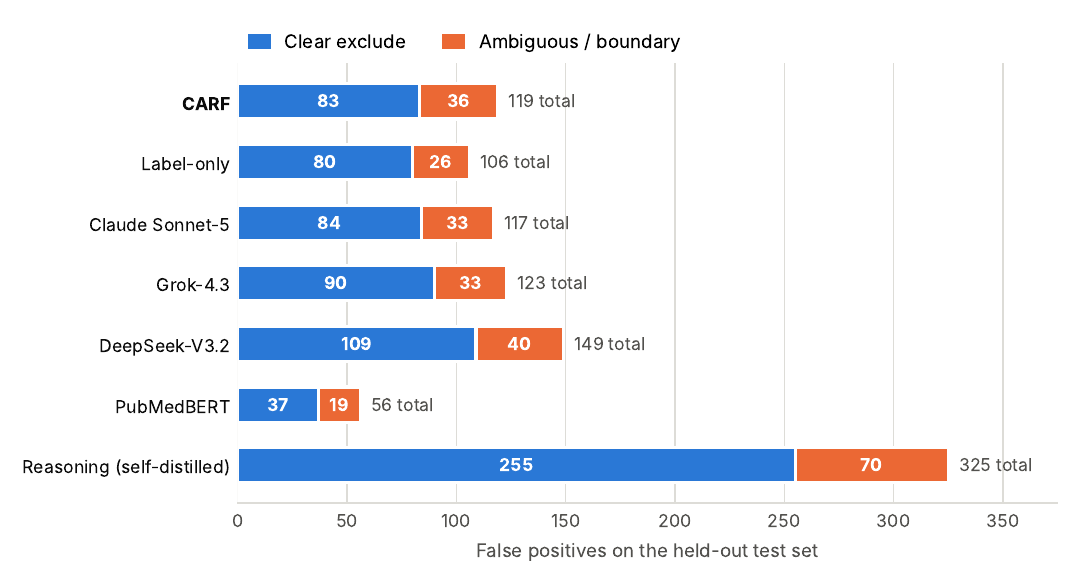}
  \caption{False positives by model, split into clearly ineligible and ambiguous articles by a blinded expert rating (single rater; 347 pooled articles).}
  \label{fig:fp-boundary}
\end{figure}

Table~\ref{tab:fp-share} reports, for each model, how many of its false positives were rated ambiguous, as a share of that model's own false positives. Among the generative models, CARF has the highest share (30.3\%), and the teacher-distilled reasoning model, trained without expert feedback, the lowest (21.5\%); the encoder baseline PubMedBERT has a higher share (33.9\%) but far fewer false positives overall (56). The differences between models are modest, and the 95\% Clopper--Pearson intervals overlap, so the ordering should be read as indicative. Figure~\ref{fig:fp-boundary} shows the same counts.

\begin{table*}[t]
\centering
\small
\setlength{\tabcolsep}{6pt}
\begin{tabular}{lcccc}
\toprule
\textbf{Model} & \textbf{FP} & \textbf{Ambig.} & \textbf{Clear} & \textbf{\% ambig. (95\% CI)} \\
\midrule
PubMedBERT                    & 56  & 19 & 37  & 33.9 (21.8--47.8) \\
CARF                          & 119 & 36 & 83  & 30.3 (22.2--39.3) \\
Claude Sonnet-5               & 117 & 33 & 84  & 28.2 (20.3--37.3) \\
Grok-4.3                      & 123 & 33 & 90  & 26.8 (19.2--35.6) \\
DeepSeek-V3.2                 & 149 & 40 & 109 & 26.8 (19.9--34.7) \\
Label-only                    & 106 & 26 & 80  & 24.5 (16.7--33.8) \\
Reasoning (teacher-distilled) & 325 & 70 & 255 & 21.5 (17.2--26.4) \\
\bottomrule
\end{tabular}
\caption{False positives of each model on the held-out test set, split by the blinded expert rating and ordered by the share rated ambiguous. Ambig.: ambiguous; Clear: clearly ineligible.}
\label{tab:fp-share}
\end{table*}

\subsection{Effect of Expert Feedback}

The teacher-distilled reasoning model shares CARF's base model, training data and reasoning format but receives no expert feedback, so comparing their false positives isolates what the feedback removes (Table~\ref{tab:fp-feedback}). Of the teacher-distilled model's clearly ineligible inclusions, 68\% are no longer included by CARF, compared with 50\% of its ambiguous inclusions (two-sided Fisher's exact test, $p=0.007$). Expert feedback therefore removes clearly ineligible inclusions more readily than borderline ones.

\begin{table}[t]
\centering
\footnotesize
\setlength{\tabcolsep}{3.5pt}
\begin{tabular}{p{0.26\columnwidth}ccc}
\toprule
\textbf{Rating} & \textbf{Teacher-dist. FP} & \textbf{Also in CARF} & \textbf{Removed} \\
\midrule
Clearly ineligible & 255 & 82 (32\%) & 173 (68\%) \\
Ambiguous          & 70  & 35 (50\%) & 35 (50\%) \\
\bottomrule
\end{tabular}
\caption{False positives of the teacher-distilled reasoning model and whether CARF also includes them.}
\label{tab:fp-feedback}
\end{table}

\section{Effect of the Amount of Expert Feedback}
\label{app:feedback-amount}
\setcounter{table}{0}
\renewcommand{\thetable}{\thesection\arabic{table}}
\setcounter{figure}{0}
\renewcommand{\thefigure}{\thesection\arabic{figure}}

\subsection{Setup}

To examine how much expert feedback CARF needs, we retrained the student model with only $N \in \{100, 500\}$ of the 1,346 expert feedback notes in the training set. Every training article was kept, with its verified screening decision; for articles whose note was removed, the training target consists of the teacher's reasoning followed directly by the verified decision, without a feedback block. The retained notes were drawn from a single fixed random ordering of all feedback articles, so the 100-note set is a subset of the 500-note set. All other settings, including data, prompt, hyperparameters, and seed, are identical to the full CARF model (Table~\ref{tab:hparams}). All three models were evaluated in the same way, on the 1000 test articles.

\subsection{Results}

\begin{table*}[t]
\centering
\small
\setlength{\tabcolsep}{6pt}
\begin{tabular}{lcccccc}
\toprule
\textbf{Feedback notes} & \textbf{Rec.} & \textbf{95\% CI} & \textbf{Spec.} & \textbf{Prec.} & \textbf{F1} & \textbf{FP} \\
\midrule
100            & 0.922 & 0.870--0.958 & 0.871 & 0.586 & 0.717 & 108 \\
500            & 0.964 & 0.923--0.987 & 0.836 & 0.539 & 0.691 & 137 \\
1,346 (full)   & 0.964 & 0.923--0.987 & 0.857 & 0.573 & 0.719 & 119 \\
\bottomrule
\end{tabular}
\caption{Screening performance of CARF trained with different amounts of expert feedback, on the 1000 test articles (166 true includes). Single training run per configuration. 95\% CI: exact Clopper--Pearson interval for recall.}
\label{tab:feedback-amount}
\end{table*}

\begin{figure}[t]
  \centering
  \includegraphics[width=0.85\columnwidth]{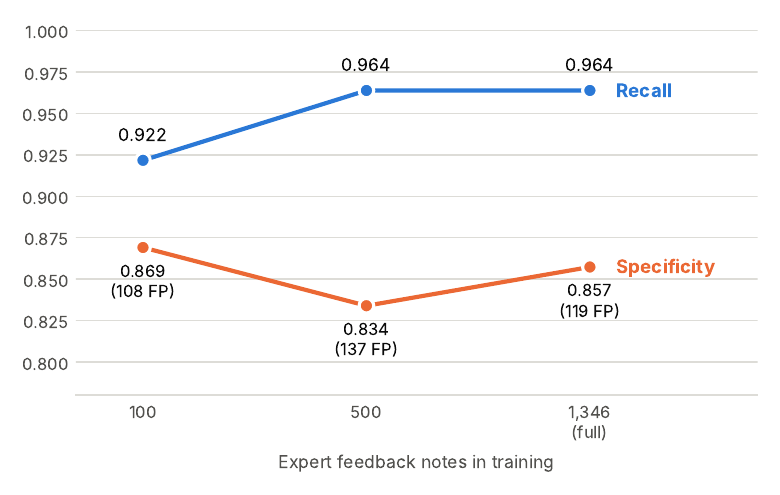}
  \caption{Include recall and specificity for CARF trained with 100, 500, and all 1,346 expert feedback notes. Specificity points are labelled with the corresponding number of false positives (FP). The full model is evaluated on all 1,000 test articles, as in Table~\ref{tab:screening-results}; Single training run per configuration.}
  \label{fig:feedback-amount}
\end{figure}

Recall increased from 0.922 with 100 notes to 0.964 with 500 notes, reducing missed eligible studies from 13 to 6. The full model did not improve on this further. False positives did not change monotonically with the amount of feedback.

We also recorded, for every test article, the probability the model assigned to an include decision. At the point where the model writes its final decision, immediately after the text ``DECISION:'', we take the model's next-token probability distribution (its softmax output, with no temperature scaling) and sum the probability of the token that begins \texttt{INCLUDE} and of the token that begins every exclusion code (\texttt{EX}), with and without a leading space. The include probability is the first sum divided by the total of the two. Because this probability is computed after the model has generated its reasoning, it reflects the model's confidence given that reasoning and is not a calibrated estimate of eligibility.

The models trained with less feedback were less certain in their decisions. For the eligible studies they missed, the include probability ranged from roughly 0.03 to 0.5. Lowering the decision threshold, so that an article is also sent to full-text review when its include probability reaches a cutoff $t$, therefore recovers missed studies more efficiently for these models: at $t = 0.2$, the 500-note model reaches a recall of 0.976 with 234 false positives.

\section{Self-Corrective Reasoning}
\label{app:self-correction}
\setcounter{figure}{0}
\renewcommand{\thefigure}{\thesection\arabic{figure}}

Analysis of the model's reasoning reveals a self-corrective behavior learned from the expert feedback during training. In several test-set examples, the adapted model initially considers evidence that could support inclusion but subsequently revisits its own evidence against the screening criteria, identifies a contextual inconsistency or eligibility violation, and corrects its initial interpretation before producing the final decision.

\begin{figure*}[t]
  \centering
  \includegraphics[width=0.42\textwidth]{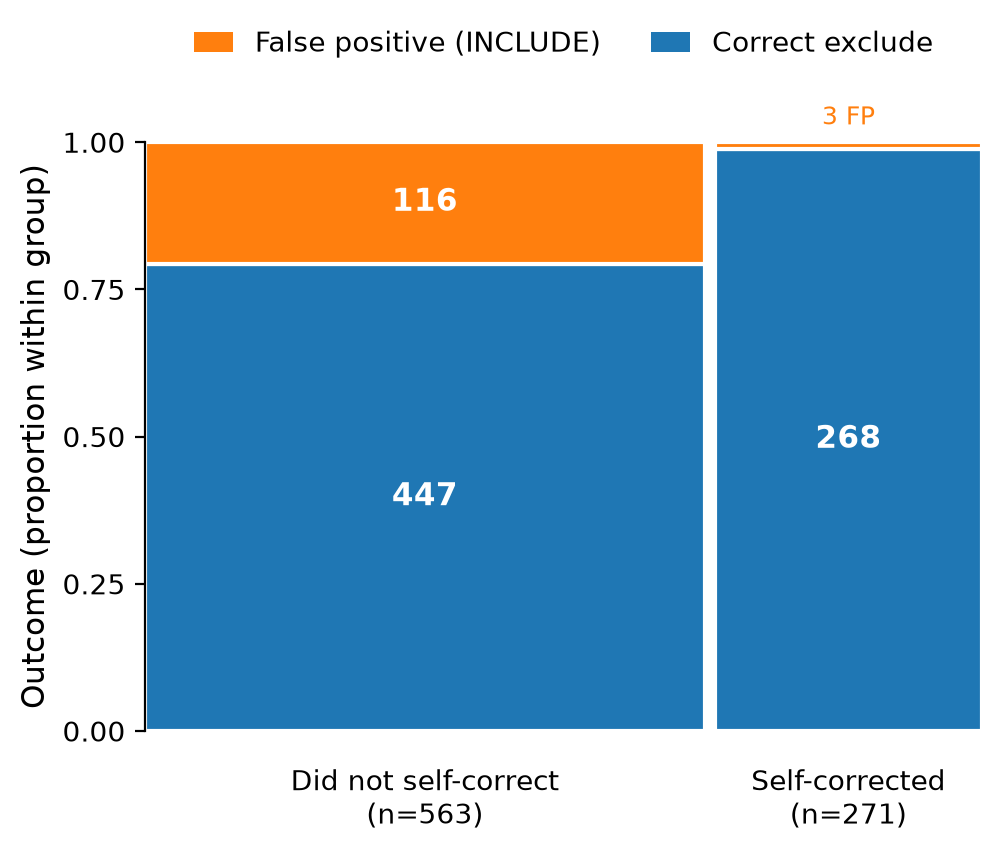}
  \caption{Screening outcomes on excluded test articles by whether the model self-corrected its reasoning.}
  \label{fig:self_correction}

  \vspace{1.2em}

  \begin{subfigure}{\textwidth}
    \centering
    \includegraphics[width=\textwidth]{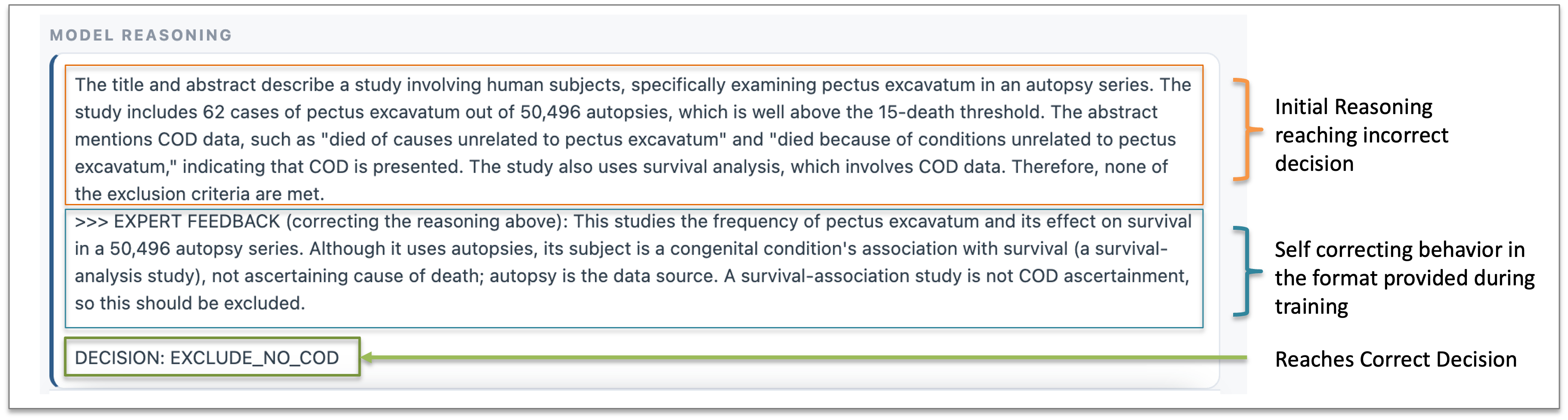}
    \caption{}
    \label{fig:qual_a}
  \end{subfigure}

  \vspace{0.75em}

  \begin{subfigure}{\textwidth}
    \centering
    \includegraphics[width=\textwidth]{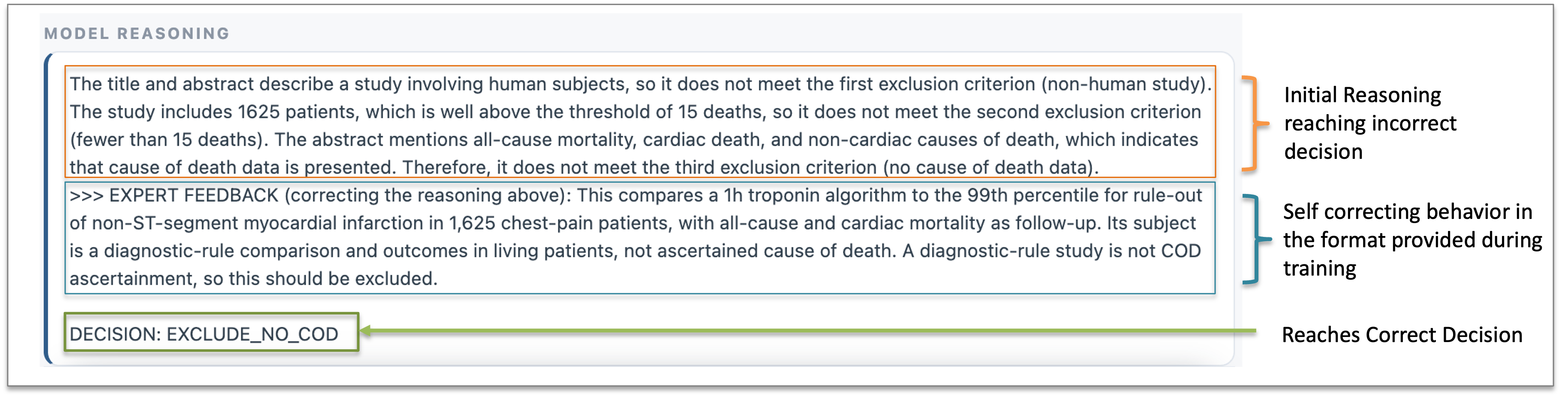}
    \caption{}
    \label{fig:qual_b}
  \end{subfigure}
  \caption{Qualitative examples of self-corrective reasoning on the held-out test set. In each case the model's initial reasoning leans toward inclusion, then the expert-feedback--informed reasoning revisits the evidence against the eligibility criteria and corrects the decision to exclude.}
  \label{fig:qualitative_examples}
\end{figure*}

\paragraph{Effect on excluded articles.} Across the 834 excluded articles in the held-out test set, the model self-corrected its reasoning on 271, of which only 3 resulted in false-positive inclusion decisions (1.1\%). Among the 563 excluded articles where the model did not self-correct, 116 resulted in false positives (20.6\%). False-positive decisions were therefore substantially less frequent when the model revisited and corrected its initial reasoning (Figure~\ref{fig:self_correction}). Figure~\ref{fig:qualitative_examples} shows two examples.

\paragraph{Effect on eligible articles.} The same behavior also accounts for most of CARF's missed studies. The model self-corrected on 11 of the 166 eligible articles, and in 4 of these the correction overturned an initially correct inclusion, leading to a false exclusion. These 4 cases make up 4 of the 6 eligible studies that CARF missed; of the 155 eligible articles on which the model did not self-correct, it missed 2.

\paragraph{Interpretation.} This pattern reflects the composition of the training data. In most expert-feedback examples, the feedback overturned an inclusion, so the model learned to associate self-correction primarily with exclusion. The behavior is therefore effective at removing ineligible studies but can also remove eligible ones. Expert feedback that also confirms correct inclusions may reduce these errors, which we leave to future work.

\section{Artifact Release and License}
\label{app:license}

We release the data artifacts produced in this work under the MIT License. These comprise the verified screening decisions, the teacher-generated reasoning traces, the expert feedback and corrections, and the fine-tuned model adapters. Article titles and abstracts were obtained from the original publishers and bibliographic databases and are not covered by this license; we distribute record identifiers so that the source texts can be retrieved under their original terms of use. The base models used in this work, DeepSeek-R1-Distill-Qwen-32B and DeepSeek-R1-Distill-Qwen-14B, are distributed under the MIT License, which permits modification and derivative works; both are derived from the Qwen-2.5 series, originally released under the Apache 2.0 License. Our use of these models is consistent with their licenses and intended research use.

\end{document}